\documentclass{bmvc2k}
\usepackage{times}
\usepackage{epsfig}
\usepackage{graphicx}
\usepackage{amsmath}
\usepackage{amssymb}
\usepackage{bm}
\usepackage{bbold}
\usepackage{gensymb}
\usepackage{multirow}
\usepackage{multicol}
\usepackage{dsfont}

\title{OMAD: Object Model with Articulated Deformations for Pose Estimation and Retrieval}

\addauthor{Han Xue$^*$}{xiaoxiaoxh@sjtu.edu.cn}{1}
\addauthor{Liu Liu$^*$}{liuliu1993@sjtu.edu.cn}{1}
\addauthor{Wenqiang Xu}{vinjohn@sjtu.edu.cn}{1}
\addauthor{Haoyuan Fu}{simon-fuhaoyuan@sjtu.edu.cn}{1}
\addauthor{Cewu Lu$^\dagger$}{lucewu@sjtu.edu.cn}{1}

\addinstitution{
 Shanghai Jiao Tong University\\
 Shanghai, CN
}

\runninghead{Xue ET AL.}{OMAD: Object Model with Articulated Deformations}

\def\eg{\emph{e.g}\bmvaOneDot}

\begin{document}

\maketitle

\begin{abstract}
Articulated objects are pervasive in daily life. However, due to the intrinsic high-DoF structure, the joint states of the articulated objects are hard to be estimated. To model articulated objects, two kinds of shape deformations namely the geometric and the pose deformation should be considered. In this work, we present a novel category-specific parametric representation called Object Model with Articulated Deformations (OMAD) to explicitly model the articulated objects. In OMAD, a category is associated with a linear shape function with shared shape basis and a non-linear joint function. Both functions can be learned from a large-scale object model dataset and fixed as category-specific priors. Then we propose an OMADNet to predict the shape parameters and the joint states from an object's single observation. With the full representation of the object shape and joint states, we can address several tasks including category-level object pose estimation and the articulated object retrieval. To evaluate these tasks, we create a synthetic dataset based on PartNet-Mobility. Extensive experiments show that our simple OMADNet can serve as a strong baseline for both tasks. Our code and dataset are available in \url{https://sites.google.com/view/omad-bmvc/}.
\end{abstract}

\section{Introduction}
\begin{figure}[t!]
\begin{center}
\includegraphics[width=0.7\linewidth]{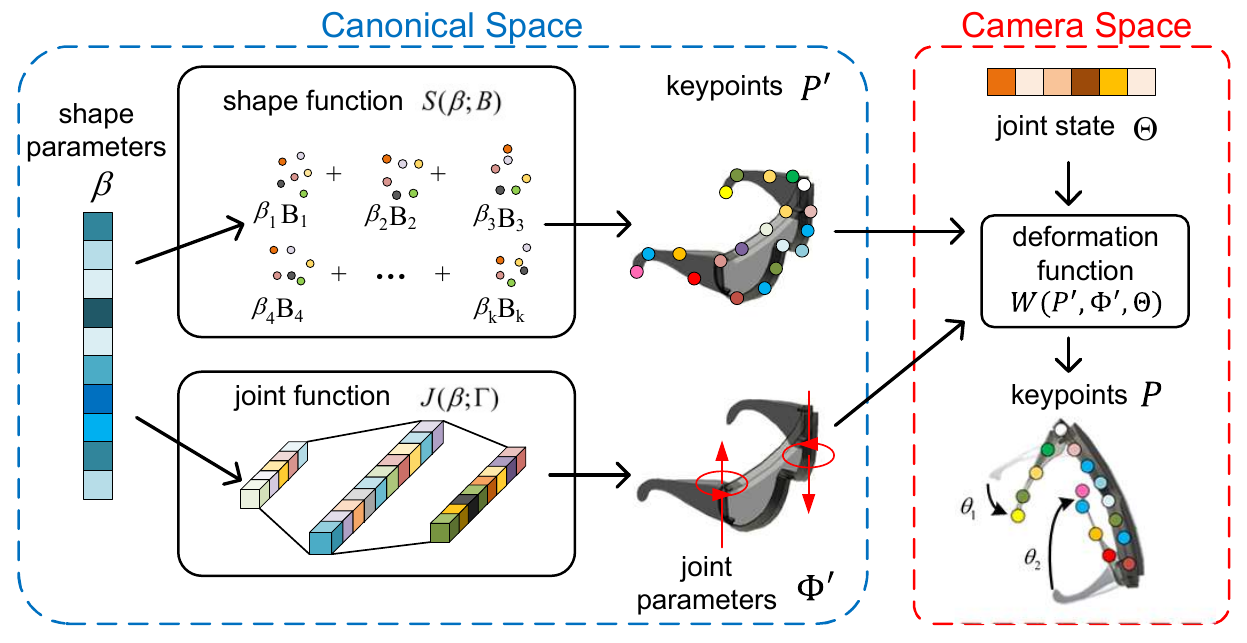}
\end{center}
   \vspace{-5mm}
   \caption{Our OMAD representation for modeling articulated objects. In OMAD, we use shape parameters $\bm{\beta}$ to capture shape geometric deformations and use joint states $\bm{\Theta}$ to capture pose deformations. We use $'$ to represent the variables in canonical space (e.g.$\bm{P'}$), which distinguishes from variables in camera space (e.g. $\bm{P}$).}
\label{fig:intro}
\vspace{-5mm}
\end{figure}

Articulated objects are very common in daily life. A vision system that can model articulated objects from a single observation (\eg depth image, point cloud) can be beneficial for the downstream tasks such as robot manipulation \cite{articulated_science_robot}, AR/VR \cite{articulated_vr} and CG modeling \cite{part_induction}. However, visual analysis of articulated objects is challenging due to their high-DoF (Degree of Freedom) nature in 3D space and large intra-class appearance variations. In this work, we propose an explicit parametric representation to depict shape and pose of articulated objects in the same category. There are two sources of articulated deformations, namely shape geometric deformation on parts and pose deformations on joints. To handle both deformations, we propose a novel \textbf{O}bject \textbf{M}odel with \textbf{A}rticulated \textbf{D}eformations (\textbf{OMAD}) (Fig. \ref{fig:intro}). 

To describe the shape geometric deformations, given an articulated object, we consider its rest structure (i.e. shape geometry, joint parameters) in the axis-aligned, zero-centered canonical space. Then for its shape, which is represented by ordered sparse keypoints, can be derived from the proposed shape function $\bm{P}'=S(\bm{\beta};\bm{B})$, where $\bm{\beta}$ is a pose-invariant shape parameters in canonical space, and $\bm{B}$ is the category-specific linear shape basis. For the joint parameters $\bm{\Phi}'$, we devise a non-linear joint function $\bm{\Phi}'=J(\bm{\beta};\bm{\Gamma})$ with parameters $\bm{\Gamma}$. The parameters $\{\bm{B}, \bm{\Gamma}\}$ of shape function and joint function can be regarded as category-specific priors, and can be learned from existing object dataset like PartNet-Mobility \cite{sapien} with proposed OMAD-PriorNet. In this way, the shape geometric deformation can be fully described by $(\bm{P}', \bm{\Phi}')$. On the other hand, the pose deformation on joints can simply defined by its joint state $\bm{\Theta}$ in camera space.

To combine both deformations, we devise a deformation function $W(\bm{P}', \bm{\Phi}', \bm{\Theta})$. It can transform the rest structure in canonical space to camera space. In this sense, we can depict arbitrary articulated object of a known category simply with $\bm{\beta}$ and $\bm{\Theta}$. To predict $\bm{\beta}$ and $\bm{\Theta}$, we propose a novel OMADNet, which is mainly constituted by a neural prediction stage for coarse estimation of $\bm{\beta}$ and $\bm{\Theta}$ and an optimization stage to refine the prediction results.
To show the applicability of the proposed OMADNet, we apply it into two different tasks, namely articulated object pose estimation and articulated object retrieval. The predicted shape parameters $\bm{\beta}$ can be used for pose-oblivious object retrieval \cite{articulate_signature_pose_oblivious}. Besides, the predicted joint state $\bm{\Theta}$ and joint parameters $\bm{\Phi}'$ can directly address the object pose estimation task. To evaluate our method, we select articulated object models of 5 categories from PartNet-Mobility\cite{sapien} to create a dataset named ArtImage. For the articulated object pose estimation task, we achieve better results on joint state error and joint distance error compared to existing method. For the object retrieval task, we achieve 82.3 average mAP on 5 categories.

Our contribution can be summarized as follows:
\begin{itemize}
    \item A novel category-specific articulated modeling approach named OMAD, which explicitly models both shape deformation and pose deformation at category-level.
    \item With the proposed OMAD, we provide a simple OMADNet to estimate the shape parameters and joint states required to describe the articulated object from its single observation. The output of OMADNet can be used to address the category-level object pose estimation task and the articulated object retrieval task.
\end{itemize}

\section{Related Work}
\noindent\textbf{Articulated Object Shape Signature and Retrieval.}
The shape parameters in our proposed OMAD which capture the  geometric features of articulated objects and summarize them into a vector are mostly like the shape signature adopted in \cite{articulate_signature1,articulate_signature2, articulate_signature3, articulate_signature4, articulate_signature_pose_oblivious, a-sdf}. However, these shape signatures cannot be directly applied to our case for at least two reasons: (1) Their methods are largely relied on the mesh representation, such as spectral-based \cite{articulate_signature1}, graph-based \cite{articulate_signature2}, geodesic-based \cite{articulate_signature_pose_oblivious}, SDF-based\cite{a-sdf}, while our shape is in point cloud format. (2) None of the previous methods are interested in the joint parameters of the articulated objects, while our representation takes both joint states and joint parameters into account. For articulated object retrieval task, the standard benchmarks are the the McGill shape benchmark (MSB) \cite{msb_dataset} and ISDB dataset \cite{articulate_signature_pose_oblivious}. However, the object models for a certain category are very few for both datasets, which cannot support the training of our proposed OMAD-PriorNet and OMADNet. 

\noindent\textbf{Low-rank Shape Prior.} Our proposed OMAD adopts a low-rank shape prior based approach to model the sparse shape. Such choice has been proved effective in many tasks, such as instance segmentation \cite{usd_seg}, NRSfM \cite{low_rank_nrsfm1,low_rank_nrsfm2,low_rank_nrsfm3,low_rank_nrsfm4}, object structure learning \cite{category_kpt}, human body modeling \cite{smpl, mano}. To the best of our knowledge, we are the first to apply it to the generic articulated object modeling. It is worth noting that our method differs from SMPL\cite{smpl} and its variants\cite{mano} from at least two aspects: (1) SMPL\cite{smpl} relies on hand-crafted mesh templates with fixed topology and vertex number, while our OMAD only requires flexible point cloud input. (2) SMPL\cite{smpl} performs PCA on standard meshes to obtain shape basis, while our proposed OMAD-PriorNet can learn shape basis in an unsupervised manner directly from raw point cloud. 

\noindent\textbf{Category-level Articulated Object Pose Estimation.} 
Cateogry-level object pose estimation task aims at predicting poses of unseen objects in the same category, which is firstly introduced by NOCS\cite{nocs}. A-NCSH\cite{ancsh} extends this task to articulated objects. Later, several works on this task are conducted on videos \cite{screwnet, art_obj_pose_model_free}. Our proposed OMADNet can address the category-level articulated object pose estimation task given a single depth image as input. Overall, our work is closely related to \cite{ancsh} with significant different on methodology. Firstly, \cite{ancsh} takes a part-centered perspective, the joint states are estimated from the predicted part segmentations and poses. Our method takes a joint-centered perspective, the joint states are directly predicted by OMADNet. Secondly, \cite{ancsh} uses dense NOCS coordinates to represent articulated objects, while our method use ordered keypoints to obtain sparse correspondence, which largely reduces optimization complexity. Thirdly, Li et al. \cite{ancsh} models the shape deformation implicitly with segmentation and NOCS coordinates, which cannot explicitly reflects the complete shape deformation by parameterization as our proposed OMAD representation. Thus it cannot support other important tasks like object retrieval.

\section{Object Model with Articulated Deformations, OMAD}\label{sec:omad}
In this section, we will give a definition about the articulated objects which the OMAD can support (Sec. \ref{sec:articulation}). As the representation of pose deformation on the joints are self-evident, we will mainly discuss how to model shape geometric deformations in canonical space(Sec. \ref{sec:OMAD_space}) with shape function (Sec. \ref{sec:shape_func}) and joint function (Sec. \ref{sec:joint_func}). A general articulated object in camera space is formed by transforming the keypoints and joint parameters in canonical space via a deformation function (Sec. \ref{sec:deform_func}).

\subsection{Object Articulation} \label{sec:articulation}
\begin{figure}[htbp]
\centering
\begin{minipage}[t]{0.68\textwidth}
\centering
\includegraphics[width=\linewidth]{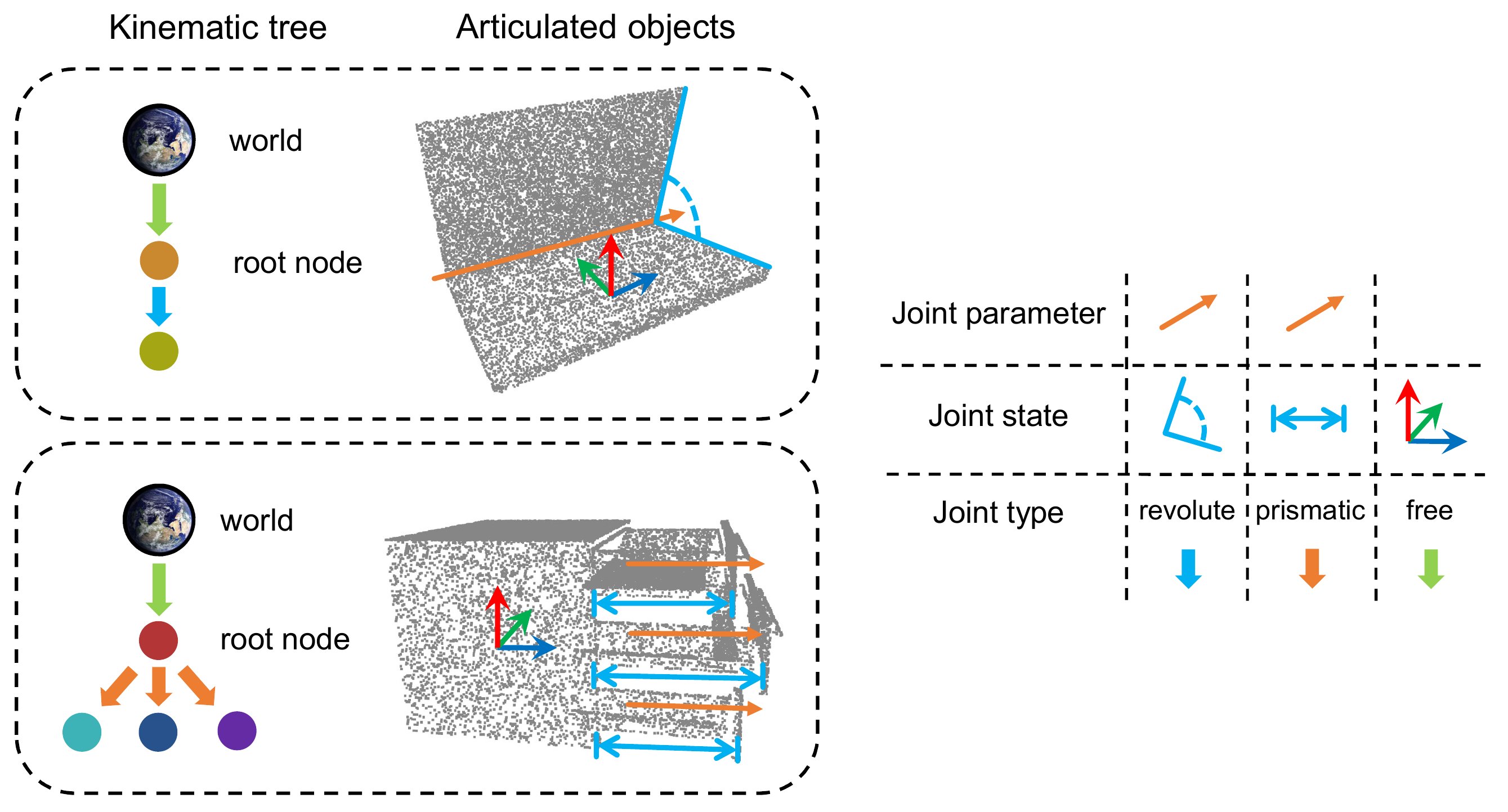}
\caption{Joint types, parameters and states in consideration. }
\label{fig:joints}
\end{minipage}
\begin{minipage}[t]{0.3\textwidth}
\centering
\includegraphics[width=0.7\linewidth]{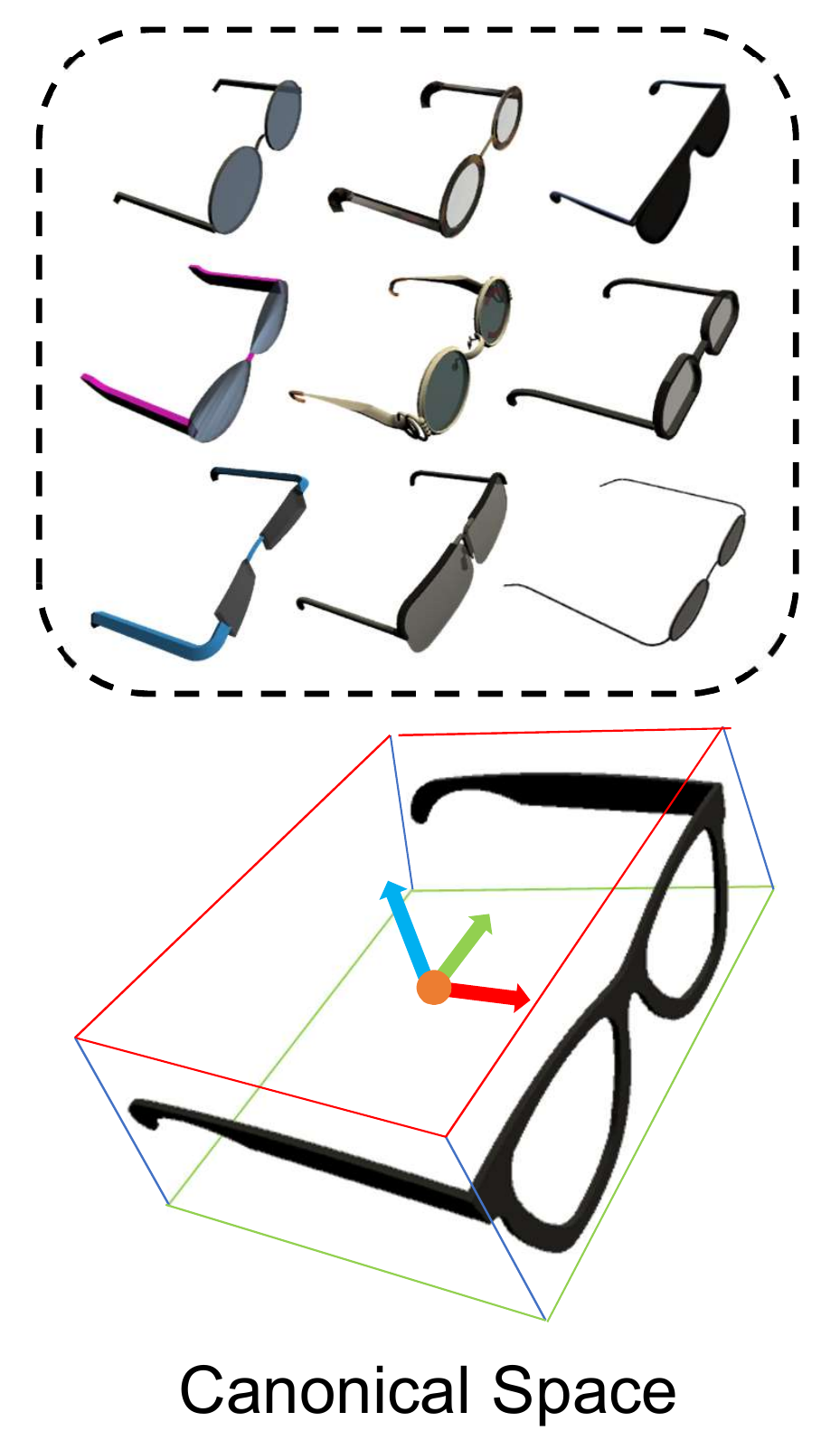}
\caption{Examples of objects in canonical space.}
\label{fig:OMAD_space}
\end{minipage}
\vspace{-4mm}
\end{figure}

We take a joint-centered perspective to study articulated objects. Kinematic tree is one kind of abstraction of articulated objects (See Fig. \ref{fig:joints}). We denote the joint that connects one node and its parent node in the kinematic tree as the \textbf{reference joint} of this node. For example, if joint $k$ connects node $a$ and its parent node $b$, then joint $k$ is the reference joint of node $a$.  

Here we consider three types of joints: free joint, revolute joint and prismatic joint (See Fig. \ref{fig:joints}). (1) Free joint is the joint that connects the world node with the base part (root node) in the kinematic tree. The joint state of free joint can be defined as $\bm{\theta}^{(free)}=(\mathbf{R}^{(f)}, \mathbf{t}^{(f)})$, which can be seen as the 6D pose of base part in the world coordinate space. (2) For revolute joint, joint parameters can be defined as $\bm{\phi}^{(r)}=(\bm{u}^{(r)},\bm{q}^{(r)})$, where $\bm{u}^{(r)}$ represents the direction of joint axis and $\bm{q}^{(r)}$ represents the pivot point of joint axis. The joint state is denoted by $\bm{\theta}^{(r)}$, which is the relative rotation angle compared to a pre-defined rest state in canonical space. (3) For prismatic joint, joint parameters can be defined as $\bm{\phi}^{p}$, where $\bm{u}^{(p)}$ represents the direction of joint axis. The joint state is denoted by $\bm{\theta}^{(p)}$, which is the relative distance along the joint axis compared to a pre-defined rest state in canonical space.

With the introduction of reference joint and free joint, the part number and the joint number can be equal. Given an articulated object with $K$ joints, the overall joint state is denoted as $\bm{\Theta}=\{\bm{\theta}_1,\ldots,\bm{\theta}_K\}$.

\subsection{Geometric Deformation in Canonical Space}\label{sec:OMAD_space}
As pose deformation can be independently considered by $\bm{\Theta}$, we model the shape geometric deformation in the canonical space (See Fig. \ref{fig:OMAD_space}). Objects of the same category in the canonical space have the following properties: (1) The directions in the joint parameters are aligned. (2) The whole object is zero-centered. (3). The joint states are the same and predefined. To note, our canonical space does not normalize the object scale, because object scale is a crucial part of the shape geometry and can be explicitly modeled by the OMAD representation. We use $'$ to represent the variables in canonical space (e.g.$\bm{P'}$), which distinguishes from variables in camera space (e.g. $\bm{P}$).
As the articulation is associated with shape and joint, thus we model the geometric deformation with a shape function and joint function.

\subsubsection{Shape Function}\label{sec:shape_func}
As detailed geometric deformation is hard to be modeled, we adopt the sparse shape representation, which seeks sparse keypoint correspondences among objects from the same category. Inspired by \cite{category_kpt}, we predict the sparse shape with the low-rank shape prior. Given a complete object point cloud in canonical space, the sparse shape $\bm{P}'$ with $M$ keypoints can be generated by shape parameters $\bm{\beta}$ using shape function $S(\cdot)$:
\begin{equation}\label{eq:shape_func}
    \bm{P}' = S(\bm{\beta};\bm{B})=\bm{B}\bm{\beta}=\sum_{i=1}^{|\bm{\beta}|}\bm{\beta}_i\bm{b}_i
\end{equation}
where $\bm{\beta}=\left[\beta_{1}, \ldots, \beta_K\right]^{T}$, and the $\bm{b}_{i} \in \mathbb{R}^{3M}$ represents linear shape basis. Let $\bm{B}=[\bm{b}_{1}, \ldots, \bm{b}_K] \in \mathbb{R}^{3M \times K}$ be the matrix of all such shape basis, which will be shared in the same category.

\subsubsection{Joint Function}\label{sec:joint_func}
Given the rest-state object in canonical space, we can define joint parameters $\bm{\phi}'=(\bm{u}', \bm{q}')$, $\bm{u}'\in\mathbb{R}^3, \bm{q}'\in\mathbb{R}^3$ for each joint. As the corresponding joint direction $\bm{u}'$ is aligned in the canonical space, and the pivot point $\bm{q}'$ for each joint are strongly correlated to the shape geometry, we propose to predict joint parameters in canonical space from shape parameters $\bm{\beta}$.
All the joint parameters  $\bm{\Phi}'=\{\bm{\phi}_1',\ldots,\bm{\phi}_K'\}$ of the $K$ joints can be generated by joint function $J(\cdot)$:
\begin{equation}\label{eq:joint_func}
    \bm{\Phi}' = J(\bm{\beta};\bm{\Gamma})
\end{equation}
where $\bm{\Gamma}$ is the parameters of the joint function $J(\cdot)$. Since the relationship between shape parameters and joint parameters in canonical space are highly non-linear, we design a two-layer MLP with ReLU activation as $J(\cdot)$.

\subsection{Deformation function}\label{sec:deform_func}
To combine both shape geometric deformation and pose deformation, we introduce a deformation function $\bm{P} = W(\bm{P}', \bm{\Phi}', \bm{\Theta})$, which can transform the sparse shape $\bm{P}'$ in zero-centered canonical space with joint parameters $\bm{\Phi}'$ according to joint states $\bm{\Theta}$ to the observed sparse shape in the camera space $\bm{P}$. Each keypoint $\bm{p}_i'$ in $\bm{P}'$ that belongs to part $k$ is transformed into $\bm{p}_i$ in $\bm{P}$ according  to Eq. \ref{eq:deform_func1} and Eq. \ref{eq:deform_func2}:
\begin{equation}\label{eq:deform_func1}
\bar{\bm{p}}_i=G_{k}(\bm{\Phi}', \bm{\Theta}) \bar{\bm{p}}_i'
\end{equation}
\begin{equation}\label{eq:deform_func2}
G_{k}(\bm{\Phi}',\bm{\Theta})=\prod_{j \in A(k)}F_j(\bm{\Phi}_j', \bm{\Theta}_j)
\end{equation}
where $\bar{\bm{p}}$ means the homogeneous form of $\bm{p}$, $G_{k}(\bm{\Phi}',\bm{\Theta})$ is the $4\times4$ transformation matrix of part $k$. Finally, $A(k)$ is the ordered set of all the ancestors of part $k$ in the kinematic tree, and $F_j(\bm{\Phi}_j',\bm{\Theta}_j)$ is the relative $4\times4$ transformation matrix caused by the reference joint of part $j$. Please refer to the supplementary materials for the details of $F_j(\bm{\Phi}_j',\bm{\Theta}_j)$.

\subsection{Learning the OMAD Prior}\label{sec:method-unsup}
The parameters of shape function $S(\bm{\beta};\bm{B})$ and joint function $J(\bm{\beta};\bm{\Gamma})$ are category-specific, and we call these parameters $\{\bm{B}, \bm{\Gamma}\}$ as \textbf{OMAD prior}. We design a network called OMAD-PriorNet to learn OMAD prior for each cateogry in an unsupervised manner.  The parameters $\{\bm{B}, \bm{\Gamma}\}$ of shape function $S(\bm{\beta};\bm{B})$ and joint function $J(\bm{\beta};\Gamma)$
will be treated as shared network parameters of OMAD-PriorNet for each category, and they will be updated during training process. After the training, these parameters will be fixed as category-specific \textbf{OMAD prior}, then we will predict shape parameters $\bm{\beta}$ and keypoints $\bm{P}'$ for each instance in the training set with the pre-trained OMAD-PriorNet which will be used as ground truth in the training of OMADNet described in Sec. \ref{coarse_prediction}.
Please refer to the supplementary materials for all the loss functions and network structure of OMAD-PriorNet. 

\noindent\textbf{Learning of Shape Function.}
To learn the shape basis $\bm{B}$ for shape function $S(\cdot)$, we adopt a similar design in \cite{category_kpt}. We make some modifications on the loss functions to adapt to the articulated objects. The tweaks on loss functions are listed in the following: (1) We add an additional $\mathcal{L}_2$ regularization loss on $\bm{\beta}$ which can improve the generalization ability. (2) The original chamfer loss, coverage loss and surface loss is changed to part-level. (3) We add a separation loss to ensure the keypoints not collapsed into a cluster. 

\noindent\textbf{Learning of Joint Function.} The parameters $\bm{\Gamma}$ of joint function $J(\cdot)$ is optimized during training according to the proposed \textbf{joint loss} defined in Eq. \ref{eq:joint_loss}.
\begin{equation}\label{eq:joint_loss}
    L_{\text{joint}} = L_{\bm{\mu}} + \lambda L_{\bm{q}}
\end{equation}
where $L_{\bm{\mu}}$ is a cosine-similarity loss that supervises joint direction and $L_{\bm{q}}$ is the $L_2$ loss between the predicted joint pivot point and the ground-truth joint pivot point.

\begin{figure*}[t!]
\centering
\begin{center}
   \includegraphics[width=0.9\linewidth]{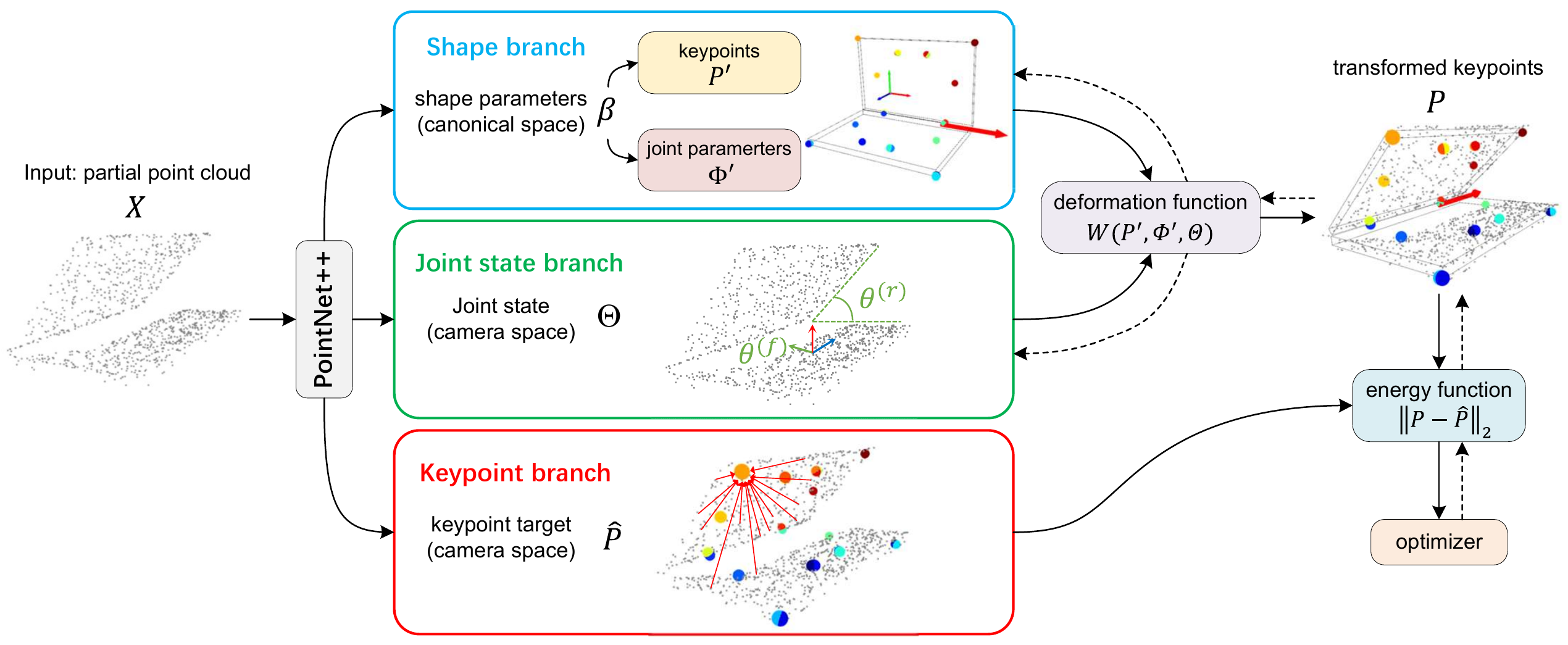}
\end{center}
   \vspace{-4mm}
   \caption{Overall pipeline of our OMADNet. Given a partial point cloud of an observed articulated object, the point features extracted by PointNet++\cite{pointnet++} will be fed into three branches: (1) shape branch (2) joint state branch (3) keypoint branch. We also devise an optimizer to refine final results.}
\label{fig:pipeline}
\vspace{-4mm}
\end{figure*}

\section{OMADNet}\label{sec:omadnet}
With the OMAD representation and the pre-learned OMAD priors, given a point cloud $\bm{X}=\left\{\bm{x}_{i} \in \mathbb{R}^{3} \mid i=1, \ldots, N\right\}$ reprojected from a depth image, we propose a OMADNet to predict the shape parameter $\bm{\beta}$ and joint states $\bm{\Theta}$. 
The overall pipeline is shown in Fig. \ref{fig:pipeline}.

\subsection{Coarse Prediction of Shape Parameters and Joint States}\label{coarse_prediction}

\noindent\textbf{Feature Extractor.} We adopt PointNet++ \cite{pointnet++} to extract $d$-dimensional dense point-wise feature $\mathcal{F}_{\text{dense}}\in \mathbb{R}^{N\times d}$ from $\bm{X}$. Then we can generate global feature $\mathcal{F}_{\text{global}}\in \mathbb{R}^{d}$ using a max pooling layer with the extracted dense feature.

\noindent\textbf{The Shape Branch.} The shape branch learns to predict pose-invariant shape parameters $\bm{\beta}$ in canonical space. It uses a two layer MLP with global feature $\mathcal{F}_{\text{global}}$ as input to regress a vector as shape parameters $\bm{\beta}$. We supervise $\bm{\beta}$ by Eq. \ref{eq:shape_loss}:
\begin{equation}\label{eq:shape_loss}
    L_{\bm{\beta}} = \left\|S(\bm{\beta})-S(\bm{\beta}^*)\right\|_2+L_{\text{joint}}(J(\bm{\beta}), J(\bm{\beta}^*)),
\end{equation}
where $\bm{\beta}^*$ is obtained by the pre-trained OMAD-PriorNet described in Sec. \ref{sec:method-unsup}, and $L_{\text{joint}}$ is defined in Eq. \ref{eq:joint_loss}. $L_{\bm{\beta}}$ indirectly supervises $\bm{\beta}$ by generating keypoints $\bm{P}'=S(\bm{\beta};\bm{B})$ and joint parameters $\bm{\Phi}'=J(\bm{\beta};\bm{\Gamma})$ in canonical space.The parameters $\{\bm{B}, \bm{\Gamma}\}$ are obtained from pre-trained OMAD-PriorNet and are fixed during training. 

\noindent\textbf{The Joint State Branch.} 
This branch predicts joint states using a two layer MLP with global feature $\mathcal{F}_{\text{global}}$ as input to regress joint states $\bm{\Theta}$.  As there are three joint types in consideration, we treat them separately. For the free joint, $\bm{\theta}_{\text{\text{base}}}=\{\bm{R}_{\text{base}}, \bm{t}_{\text{base}}\}$, we directly supervise $\bm{R}_{\text{base}}$, which is represented by a quaternion vector, via a cosine-similarity loss $\mathcal{L}_{base(R)}$ while the loss $\mathcal{L}_{base(t)}$ for $\bm{t}_{\text{base}}$ is a Root Mean Square (RMS) loss. For the 1D revolute ($\mathcal{L}_r$) or prismatic ($\mathcal{L}_p$) joint, we use $L_2$ loss on $\bm{\theta}^{(r)}$ or $\bm{\theta}^{(p)}$. The overall loss for joint state is:
\begin{equation}
    L_{\Theta} = \sum_{j=1}^K \mathbb{1}^{(free)}_j(\mathcal{L}_{base(R)} + \mathcal{L}_{base(t)}) + \mathbb{1}^{(r)}_j\mathcal{L}_r + \mathbb{1}^{(p)}_j\mathcal{L}_p,
\end{equation}
where $\mathbb{1}^{(\cdot)}_j$ indicates whether it is the joint type for the joint $j$.

\noindent\textbf{The Keypoint Branch.} The keypoint branch will directly predict ordered keypoint targets $\hat{\bm{P}}\in\mathbb{R}^{M\times 3}$ in camera space using dense feature $\mathcal{F}_{\text{dense}}$. These keypoints are for the optimization process described later in Sec. \ref{sec:optimizer}. Similar to PVN3D \cite{pvn3d}, we predict $\hat{\bm{P}}$ with an offset-voting mechanism, that is every point will regress all the $M$ keypoint offsets with respect to its own location. These offsets are denoted by $\mathcal{O}\in\mathbb{R}^{N\times M}$. However, it is hard to directly regress keypoint offsets in camera space for articulated objects due to large joint deformations and self-occlusion. So we additionally predict an attention map $\mathcal{S}$ of input point cloud $\bm{X}$ to focus areas around keypoints. Specifically, the $j$-th keypoint $\hat{\bm{p}}_j\in\hat{\bm{P}}$ is generated by Eq. \ref{eq-sup_kp}:
\begin{equation}\label{eq-sup_kp}
    \hat{\bm{p}}_j = \frac{1}{N}\sum_{i=1}^{N}s_{i,j}(\bm{x}_i + \bm{o}_{i, j})
\end{equation}
where $s_{i,j}\in[0, 1]$ is the attention score of $i$-th point for $j$-th keypoint,  $\bm{o}_{i,k} \in \mathcal{O}$ is the 3D offset of $j$-th keypoint repective to the $i$-th point, $\bm{x}_i \in \bm{X}$ is the 3D coordinates of $i$-th input point. The prediction of $\hat{\bm{P}}$ is supervised by:
\begin{equation}
    L_{\text{kp}}=\left\|\hat{\bm{P}} - W(S(\bm{\beta}^*;\bm{B}), J(\bm{\beta}^*;\bm{\Gamma}), \bm{\Theta}^*)\right\|_2
\end{equation}
where $\bm{\Theta}^*$ is the ground truth of the joint state, the rest of deformation function $W(\cdot)$ is obtained from pre-trained OMAD-PriorNet in Sec. \ref{sec:method-unsup}.

\subsection{Optimization-based Estimator}\label{sec:optimizer}
Since the regression of neural network is known to be noisy, empirically the estimated high-level properties like shape parameters $\bm{\beta}$ and joint states $\bm{\Theta}$ are less accurate than the predicted ordered keypoint targets $\hat{\bm{P}}$ in camera space. Thus, to push the limits of the prediction results in inference, we propose to refine $\bm{\beta}$ and $\bm{\Theta}$ with an energy function $E$ defined in Eq. \ref{eq:optimization}:
\begin{equation}\label{eq:optimization}
\begin{split}
    E & = \mathop{\arg\min}_{\bm{\beta}, \bm{\Theta}}\left\|W(S(\bm{\beta};\bm{B}), J(\bm{\beta};\bm{\Gamma}), \bm{\Theta}) -\hat{\bm{P}}\right\|_2 \\
    &=\mathop{\arg\min}_{\bm{\beta}, \bm{\Theta}}\left\|\bm{P} -\hat{\bm{P}}\right\|_2
\end{split}
\end{equation}
where $\bm{P}$ is the camera-space keypoints generated by deformation function $W(\cdot)$. The energy function $E$ is differentiable, so we can try to find optimal shape parameters $\bm{\beta}$ and joint states $\bm{\Theta}$ to fit the predicted keypoint targets $\hat{\bm{P}}$ during the optimization process. The initial values of $\bm{\beta}$ and $\bm{\Theta}$ are obtained from OMADNet prediction, which vastly reduces the optimization time. Please refer to the supplementary files for further details.

\section{Experiments}\label{sec:exp}
\subsection{Dataset}
To the best of our knowledge, there is no public image dataset for articulated object tasks in category level. So we create a synthetic dataset named \textbf{ArtImage} to evaluate OMADNet. We select 5 categories of object models from PartNet-Mobility dataset\cite{sapien} and render depth images under random joint states, object positions and viewpoints for each category with Unity\cite{unity}. We will make our dataset public in the future. Please refer to the supplementary materials for further details of ArtImage. 

\subsection{Experiments on Category-level Pose Estimation}
\textbf{Metrics.} We use the following metrics to evaluate the performance of our algorithm on category-level pose estimation. (1) \textbf{joint state}. All the joint states error are evaluated in camera space, including mean angle error for revolute joints (max error $90\degree$ for missing ones) and mean distance error for prismatic joints  (max error $1$ for missing ones). (2) \textbf{joint parameters}. All the joint parameters are evaluated in camera space. we use mean angle error to evaluate the orientation of joint axis, and use mean distance error to evaluate the distance between two joint axes.

\begin{table*}[!thb]

\centering
\caption{Performance comparison on category-level pose estimation task on ArtImage. \textit{OMAD(initial)} represents the initial joint state and joint parameters predicted by OMADNet, and \textit{OMAD(refine)} represents the refined results with our optimizer. }
\resizebox{0.9\linewidth}{!}{
\begin{tabular}{c|c|c|c|c|c}
\hline
\multirow{2}{*}{Category} & \multirow{2}{*}{Method} & \multicolumn{1}{c|}{Joint State} & \multicolumn{2}{c|}{Joint Parameter} & \multirow{2}{*}{Inference Time per Image $\downarrow$}\\
\cline{3-5}
& & error $\downarrow$ & angle error $\downarrow$ & distance error $\downarrow$\\
\hline
\multirow{3}{*}{Laptop} & A-NCSH\cite{ancsh} & 3.5\degree & \textbf{1.7\degree} & 0.09 & 9.0s \\
& OMAD(initial) & 5.9\degree & 19.5\degree & 0.14 & \textbf{0.4s} \\
 & OMAD(refine) & \textbf{3.3}\degree & 3.7\degree & \textbf{0.03} & 1.6s \\
\hline
\multirow{3}{*}{Eyeglasses} & A-NCSH\cite{ancsh} &  12.8\degree, 14.2\degree & \textbf{3.1\degree}, \textbf{3.1\degree} & 0.07, 0.06 & 11.9s \\

& OMAD(initial) &  19.1\degree, 19.4\degree & 17.3\degree, 17.3\degree & 0.18, 0.17 & \textbf{0.7s} \\

 & OMAD(refine) & \textbf{4.9\degree}, \textbf{5.2\degree} & 4.2\degree, 4.6\degree & \textbf{0.05}, \textbf{0.04} &  2.5s \\
 
\hline
\multirow{3}{*}{Dishwasher} & A-NCSH\cite{ancsh} & 3.8\degree & 6.1\degree & 0.11 & 5.5s \\

& OMAD(initial) & 17.6\degree & 15.7\degree & 0.17 & \textbf{0.6s} \\

 & OMAD(refine) & \textbf{3.7}\degree & \textbf{4.6\degree} & \textbf{0.09} &  1.6s \\
\hline
\multirow{3}{*}{Scissors} & A-NCSH\cite{ancsh} &  4.4\degree & \textbf{0.8\degree} & \textbf{0.04} & 6.5s \\
& OMAD(initial) &  5.8\degree & 18.7\degree & 0.14 & \textbf{0.7s} \\

 & OMAD(refine) & \textbf{3.2\degree} & 2.7\degree & 0.05 & 1.7s \\
\hline

\multirow{3}{*}{Drawer} & A-NCSH\cite{ancsh} & 0.38, 0.45, 0.41 & \textbf{2.6\degree, 2.7\degree}, 5.2 & - & 16.5s\\
& OMAD(initial) & 0.18, 0.17, 0.18 & 15.2\degree, 15.2\degree, 15.2\degree & - & \textbf{1.0s} \\

 & OMAD(refine) & \textbf{0.11, 0.11, 0.09} & 3.2\degree, 3.2\degree, \textbf{3.2\degree} & - & 1.9s \\
 \hline
\end{tabular}\label{tab:main-performance}
}
\vspace{-4mm}
\end{table*}

\begin{figure}[hb!]
\begin{center}
  \includegraphics[width=0.7\linewidth]{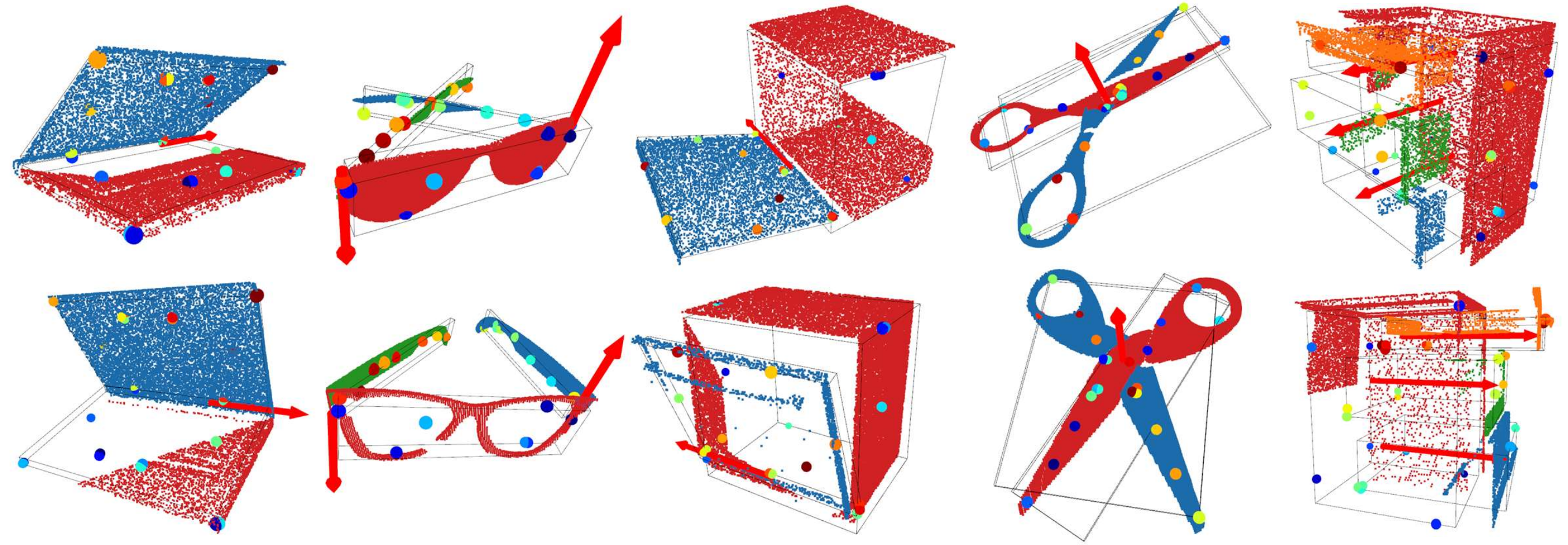}
\end{center}
\vspace{-4mm}
   \caption{Qualitative results of OMADNet on \textbf{unseen} instances for category-level object pose estimation. Keypoint indexes are represented by colors for each category. Ground-truth part segmentation are used for a better view.}
\vspace{-4mm}
\label{fig:vis_joint}
\end{figure}

\noindent\textbf{Main Results.} Table \ref{tab:main-performance} shows the performance of different methods on category-level pose estimation. We can see that OMAD(refine) surpasses OMAD(initial) by a large margin, which proves the effectiveness of our optimization-based estimator. Compared to state-of-the-art method A-NCSH\cite{ancsh}, our OMAD(refine) achieves better results on joint state error and joint distance error on most categories. It is worth noting that the prediction of A-NCSH\cite{ancsh} heavily relies on part segmentation, which could result in missing parts or joints when segmentation fails. While our method could utilize kinematic prior information to prevent missing joints. Besides, our method is much faster than A-NCSH because we adopt sparse keypoints for optimization. We show some qualitative results of our OMAD(refine) in Fig. \ref{fig:vis_joint}. We can see that the predicted keypoints after optimization can achieve semantic consistency in category, and maintain object structure under heavy self-occlusion.

\noindent\textbf{Ablation Study.} Table \ref{tab:ablation} shows some ablation experiment results. We use optimization-based estimator as default setting in all the ablation experiments. We can see from Table \ref{tab:ablation} that the use of attention score vastly boosts the performance on all metrics when predicting keypoint targets. We also analyze the upper bound of our method in Table \ref{tab:ablation}. The overall performance improves a lot when we replace keypoint targets $\hat{\bm{P}}$ predicted by keypoint branch with ground-truth keypoints during optimization process. This indicates that our OMAD representation has great potential and the bottleneck of our method is the quality of predicted keypoint locations in camera space. 
Please see supplementary files for ablation study on keypoint number, shape parameters $\bm{\beta}$ and experiment results on real images.

\begin{table}[!thb]
\centering
\caption{The effects of using attention score and ground-truth(GT) keypoint targets in OMADNet. Optimization-based estimator are used as default setting.}
\resizebox{0.8\linewidth}{!}{
\begin{tabular}{c|c|c|c|c|c}
\hline
\multirow{2}{*}{Category} & \multicolumn{2}{c|}{Ablation} & \multicolumn{1}{c|}{Joint State} & \multicolumn{2}{c}{Joint Parameter} \\
\cline{2-6}
& attention score & GT keypoints & error $\downarrow$ & angle error $\downarrow$ & distance error $\downarrow$\\
\hline

\multirow{3}{*}{Laptop} & & & 4.7\degree & 5.4\degree  & 0.04\\

 &\checkmark & & 3.3\degree & 3.7\degree & 0.03 \\

&  & \checkmark& \textbf{1.1\degree} & \textbf{0.8\degree} & \textbf{0.01} \\
\hline

\multirow{3}{*}{Eyeglasses} & & & 9.8\degree, 11.2\degree & 9.2\degree, 9.2\degree & 0.10, 0.09 \\

 & \checkmark& & 4.9\degree, 5.2\degree & 4.2\degree, 4.6\degree & 0.05, 0.04 \\
 
 & & \checkmark& \textbf{1.1\degree}, \textbf{1.2\degree} & \textbf{1.2\degree, 2.1\degree} & \textbf{0.02}, \textbf{0.01} \\
 \hline
 
 \multirow{3}{*}{Dishwasher} & & & 5.0\degree & 5.8\degree & 0.10 \\

 & \checkmark& & 3.7\degree & 4.6\degree & 0.09 \\

& & \checkmark& \textbf{0.4\degree} & \textbf{0.4\degree} & \textbf{0.02}\\
\hline

 \multirow{3}{*}{Scissors} & & & 3.9\degree & 3.9\degree & 0.05\\

 &\checkmark & & 3.2\degree & 2.7\degree & 0.05 \\

& & \checkmark& \textbf{0.4\degree} & \textbf{1.8\degree} & \textbf{0.01} \\
\hline

\multirow{3}{*}{Drawer} & & & 0.173, 0.187, 0.181 & 15.2\degree, 15.2\degree, 15.2\degree & - \\

 & \checkmark& & 0.110, 0.114, 0.093 & 3.2\degree, 3.2\degree, 3.2\degree & - \\

& & \checkmark & \textbf{0.001, 0.001, 0.003} & \textbf{0.5\degree, 0.5\degree, 0.5\degree} & - \\
\hline
\end{tabular}}\label{tab:ablation}
\vspace{-4mm}
\end{table}

\subsection{Experiments on Articulated Object Retrieval}
\textbf{Metrics.} The pose-invariant shape parameters predicted by OMADNet can be used as shape signatures for articulated object retrieval task. We use the Euclidean distance of shape parameters to estimate the similarity between object instances. We randomly choose $100$ depth images as test query set for each category. We report the mAP (mean Average Precision) of top-$10$ instances on five categories in the test dataset.

\begin{figure}[ht!]
\begin{center}
  \includegraphics[width=0.4\linewidth]{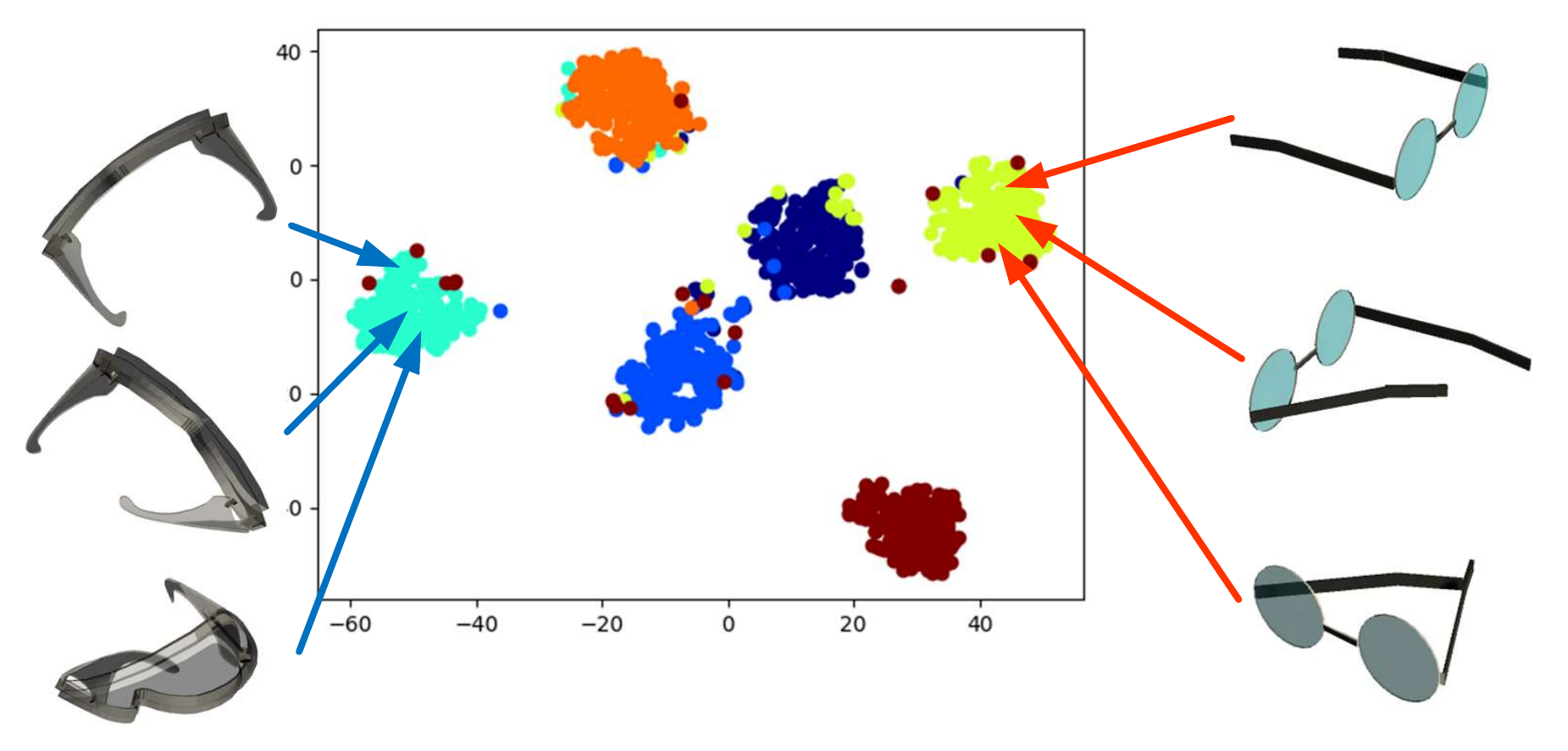}
\end{center}
\vspace{-4mm}
   \caption{The t-SNE\cite{t-SNE} visualization of predicted shape parameters of \textbf{unseen} eyeglasses instances under different poses on test dataset of ArtImage. Different instances are distinguished by colors.}
\label{fig:vis_retrieval}
\vspace{-4mm}
\end{figure}

\noindent\textbf{Results.}
To the best of our knowledge, there is no available retrieval method specially designed for our setting. Inspired from \cite{siamese}, we design a simple but effective Siamese Network as baseline which uses PointNet++\cite{pointnet++} backbone and contrastive loss\cite{contrastive} for training. Please see supplementary files for further details of the Siamese Network. Table \ref{tab:retrieval} shows some quantitative results. Although our method are not specially designed for retrieval task, we still achieve better performance compared to the baseline on most categories. We find that the Siamese Network can not distinguish some similar instances in drawer category, which vastly harms the performance, while our method can distinguish them very well.
Fig. \ref{fig:vis_retrieval} shows t-SNE\cite{t-SNE} visualization of shape parameters predicted by OMADNet. We can see that the predicted shape parameters can distinguish different instances well and maintain pose-invariant. 

\begin{table}[!thb]
\vspace{-2mm}
\centering
\caption{Articulated object retrieval results on ArtImage.}
\resizebox{0.6\linewidth}{!}{
\begin{tabular}{c|c|c|c|c|c||c}
\hline
\multirow{2}{*}{Method} & \multicolumn{6}{c}{mAP} \\
\cline{2-7}
& Laptop & Eyeglasses & Dishwasher & Scissors & Drawer & Mean \\ 
\hline
\textbf{OMAD(Ours)} & \textbf{45.2} & 93.5 & \textbf{77.7} & \textbf{98.5} & \textbf{96.7} & \textbf{82.3} \\
\textbf{Siamese Network} & 43.9 & \textbf{96.2} & 75.7 & 96.8 & 73.6 & 77.2 \\
\hline
\end{tabular}}\label{tab:retrieval}
\vspace{-4mm}
\end{table}

\section{Conclusion}
In this work, we propose a novel category-level representation, OMAD, for articulated object by modeling the two kinds of deformations namely the shape geometric deformation and pose deformation. The OMAD can describe articulated objects with shape parameters and joint states. Then for a naturally observed object, we propose to predict such OMAD information via OMADNet. The predicted OMAD information can be used for multiple tasks, such as category-level object pose estimation and articulated object retrieval.

\bibliography{egbib}

\begin{thebibliography}{29}
\providecommand{\natexlab}[1]{#1}
\providecommand{\url}[1]{\texttt{#1}}
\expandafter\ifx\csname urlstyle\endcsname\relax
  \providecommand{\doi}[1]{doi: #1}\else
  \providecommand{\doi}{doi: \begingroup \urlstyle{rm}\Url}\fi

\bibitem[uni()]{unity}
{Unity game engine.}
\newblock \url{http://www.unity.com/}.

\bibitem[Agathos et~al.(2010)Agathos, Pratikakis, Papadakis, Perantonis,
  Azariadis, and Sapidis]{articulate_signature2}
Alexander Agathos, Ioannis Pratikakis, Panagiotis Papadakis, Stavros
  Perantonis, Philip Azariadis, and Nickolas~S Sapidis.
\newblock 3d articulated object retrieval using a graph-based representation.
\newblock \emph{The Visual Computer}, 26\penalty0 (10):\penalty0 1301--1319,
  2010.

\bibitem[Akhter et~al.(2008)Akhter, Sheikh, Khan, and Kanade]{low_rank_nrsfm3}
Ijaz Akhter, Yaser Sheikh, Sohaib Khan, and Takeo Kanade.
\newblock Nonrigid structure from motion in trajectory space.
\newblock 2008.

\bibitem[Bregler et~al.(2000)Bregler, Hertzmann, and Biermann]{low_rank_nrsfm1}
Christoph Bregler, Aaron Hertzmann, and Henning Biermann.
\newblock Recovering non-rigid 3d shape from image streams.
\newblock In \emph{Proceedings IEEE Conference on Computer Vision and Pattern
  Recognition. CVPR 2000 (Cat. No. PR00662)}, volume~2, pages 690--696. IEEE,
  2000.

\bibitem[Chopra et~al.(2005)Chopra, Hadsell, and LeCun]{siamese}
Sumit Chopra, Raia Hadsell, and Yann LeCun.
\newblock Learning a similarity metric discriminatively, with application to
  face verification.
\newblock In \emph{2005 IEEE Computer Society Conference on Computer Vision and
  Pattern Recognition (CVPR'05)}, volume~1, pages 539--546. IEEE, 2005.

\bibitem[Desingh et~al.(2019)Desingh, Lu, Opipari, and
  Jenkins]{articulated_science_robot}
Karthik Desingh, Shiyang Lu, Anthony Opipari, and Odest~Chadwicke Jenkins.
\newblock Efficient nonparametric belief propagation for pose estimation and
  manipulation of articulated objects.
\newblock \emph{Science Robotics}, 4\penalty0 (30), 2019.

\bibitem[Fernandez-Labrador et~al.(2020)Fernandez-Labrador, Chhatkuli, Paudel,
  Guerrero, Demonceaux, and Gool]{category_kpt}
Clara Fernandez-Labrador, Ajad Chhatkuli, Danda Paudel, Jose Guerrero,
  C{\'e}dric Demonceaux, and Luc Gool.
\newblock Unsupervised learning of category-specific symmetric 3d keypoints
  from point sets.
\newblock In \emph{16TH EUROPEAN CONFERENCE ON COMPUTER VISION, ECCV 2020}.
  Springer, 2020.

\bibitem[Gal et~al.(2007)Gal, Shamir, and
  Cohen-Or]{articulate_signature_pose_oblivious}
Ran Gal, Ariel Shamir, and Daniel Cohen-Or.
\newblock Pose-oblivious shape signature.
\newblock \emph{IEEE transactions on visualization and computer graphics},
  13\penalty0 (2):\penalty0 261--271, 2007.

\bibitem[Hadsell et~al.(2006)Hadsell, Chopra, and LeCun]{contrastive}
Raia Hadsell, Sumit Chopra, and Yann LeCun.
\newblock Dimensionality reduction by learning an invariant mapping.
\newblock In \emph{2006 IEEE Computer Society Conference on Computer Vision and
  Pattern Recognition (CVPR'06)}, volume~2, pages 1735--1742. IEEE, 2006.

\bibitem[He et~al.(2020)He, Sun, Huang, Liu, Fan, and Sun]{pvn3d}
Yisheng He, Wei Sun, Haibin Huang, Jianran Liu, Haoqiang Fan, and Jian Sun.
\newblock Pvn3d: A deep point-wise 3d keypoints voting network for 6dof pose
  estimation.
\newblock In \emph{Proceedings of the IEEE/CVF conference on computer vision
  and pattern recognition}, pages 11632--11641, 2020.

\bibitem[Jain et~al.(2020)Jain, Lioutikov, and Niekum]{screwnet}
Ajinkya Jain, Rudolf Lioutikov, and Scott Niekum.
\newblock Screwnet: Category-independent articulation model estimation from
  depth images using screw theory.
\newblock \emph{arXiv preprint arXiv:2008.10518}, 2020.

\bibitem[Jain and Zhang(2007)]{articulate_signature1}
Varun Jain and Hao Zhang.
\newblock A spectral approach to shape-based retrieval of articulated 3d
  models.
\newblock \emph{Computer-Aided Design}, 39\penalty0 (5):\penalty0 398--407,
  2007.

\bibitem[Kim et~al.(2004)Kim, Park, Yun, and Lee]{articulate_signature3}
Duck~Hoon Kim, In~Kyu Park, Il~Dong Yun, and Sang~Uk Lee.
\newblock A new mpeg-7 standard: Perceptual 3-d shape descriptor.
\newblock In \emph{Pacific-Rim Conference on Multimedia}, pages 238--245.
  Springer, 2004.

\bibitem[Kong and Lucey(2019)]{low_rank_nrsfm4}
Chen Kong and Simon Lucey.
\newblock Deep non-rigid structure from motion.
\newblock In \emph{Proceedings of the IEEE/CVF International Conference on
  Computer Vision}, pages 1558--1567, 2019.

\bibitem[Li et~al.(2020)Li, Wang, Yi, Guibas, Abbott, and Song]{ancsh}
Xiaolong Li, He~Wang, Li~Yi, Leonidas~J Guibas, A~Lynn Abbott, and Shuran Song.
\newblock Category-level articulated object pose estimation.
\newblock In \emph{Proceedings of the IEEE/CVF Conference on Computer Vision
  and Pattern Recognition}, pages 3706--3715, 2020.

\bibitem[Liu et~al.(2020)Liu, Qiu, Wang, Hager, and
  Yuille]{art_obj_pose_model_free}
Qihao Liu, Weichao Qiu, Weiyao Wang, Gregory~D Hager, and Alan~L Yuille.
\newblock Nothing but geometric constraints: A model-free method for
  articulated object pose estimation.
\newblock \emph{arXiv preprint arXiv:2012.00088}, 2020.

\bibitem[Loper et~al.(2015)Loper, Mahmood, Romero, Pons-Moll, and Black]{smpl}
Matthew Loper, Naureen Mahmood, Javier Romero, Gerard Pons-Moll, and Michael~J
  Black.
\newblock Smpl: A skinned multi-person linear model.
\newblock \emph{ACM transactions on graphics (TOG)}, 34\penalty0 (6):\penalty0
  1--16, 2015.

\bibitem[Mu et~al.(2021)Mu, Qiu, Kortylewski, Yuille, Vasconcelos, and
  Wang]{a-sdf}
Jiteng Mu, Weichao Qiu, Adam Kortylewski, Alan Yuille, Nuno Vasconcelos, and
  Xiaolong Wang.
\newblock A-sdf: Learning disentangled signed distance functions for
  articulated shape representation.
\newblock \emph{arXiv preprint arXiv:2104.07645}, 2021.

\bibitem[Papadakis et~al.(2008)Papadakis, Pratikakis, Theoharis, Passalis, and
  Perantonis]{articulate_signature4}
Panagiotis Papadakis, Ioannis Pratikakis, Theoharis Theoharis, Georgios
  Passalis, and Stavros Perantonis.
\newblock 3d object retrieval using an efficient and compact hybrid shape
  descriptor.
\newblock In \emph{Eurographics Workshop on 3D object retrieval}, 2008.

\bibitem[Qi et~al.(2017)Qi, Yi, Su, and Guibas]{pointnet++}
Charles~R Qi, Li~Yi, Hao Su, and Leonidas~J Guibas.
\newblock Pointnet++: Deep hierarchical feature learning on point sets in a
  metric space.
\newblock \emph{arXiv preprint arXiv:1706.02413}, 2017.

\bibitem[Romero et~al.(2017)Romero, Tzionas, and Black]{mano}
Javier Romero, Dimitrios Tzionas, and Michael~J Black.
\newblock Embodied hands: Modeling and capturing hands and bodies together.
\newblock \emph{ACM Transactions on Graphics (ToG)}, 36\penalty0 (6):\penalty0
  1--17, 2017.

\bibitem[Siddiqi et~al.(2008)Siddiqi, Zhang, Macrini, Shokoufandeh, Bouix, and
  Dickinson]{msb_dataset}
Kaleem Siddiqi, Juan Zhang, Diego Macrini, Ali Shokoufandeh, Sylvain Bouix, and
  Sven Dickinson.
\newblock Retrieving articulated 3-d models using medial surfaces.
\newblock \emph{Machine vision and applications}, 19\penalty0 (4):\penalty0
  261--275, 2008.

\bibitem[Tang et~al.(2020)Tang, Xu, Ye, Yang, and Lu]{usd_seg}
Tutian Tang, Wenqiang Xu, Ruolin Ye, Lixin Yang, and Cewu Lu.
\newblock Learning universal shape dictionary for realtime instance
  segmentation.
\newblock \emph{arXiv preprint arXiv:2012.01050}, 2020.

\bibitem[Taylor et~al.(2020)Taylor, McNicholas, and Cosker]{articulated_vr}
Catherine Taylor, Robin McNicholas, and Darren Cosker.
\newblock Transporting real world rigid and articulated objects into egocentric
  vr experiences.
\newblock In \emph{2020 IEEE Conference on Virtual Reality and 3D User
  Interfaces Abstracts and Workshops (VRW)}, pages 623--624. IEEE, 2020.

\bibitem[Torresani et~al.(2008)Torresani, Hertzmann, and
  Bregler]{low_rank_nrsfm2}
Lorenzo Torresani, Aaron Hertzmann, and Chris Bregler.
\newblock Nonrigid structure-from-motion: Estimating shape and motion with
  hierarchical priors.
\newblock \emph{IEEE transactions on pattern analysis and machine
  intelligence}, 30\penalty0 (5):\penalty0 878--892, 2008.

\bibitem[Van~der Maaten and Hinton(2008)]{t-SNE}
Laurens Van~der Maaten and Geoffrey Hinton.
\newblock Visualizing data using t-sne.
\newblock \emph{Journal of machine learning research}, 9\penalty0 (11), 2008.

\bibitem[Wang et~al.(2019)Wang, Sridhar, Huang, Valentin, Song, and
  Guibas]{nocs}
He~Wang, Srinath Sridhar, Jingwei Huang, Julien Valentin, Shuran Song, and
  Leonidas~J Guibas.
\newblock Normalized object coordinate space for category-level 6d object pose
  and size estimation.
\newblock In \emph{Proceedings of the IEEE/CVF Conference on Computer Vision
  and Pattern Recognition}, pages 2642--2651, 2019.

\bibitem[Xiang et~al.(2020)Xiang, Qin, Mo, Xia, Zhu, Liu, Liu, Jiang, Yuan,
  Wang, et~al.]{sapien}
Fanbo Xiang, Yuzhe Qin, Kaichun Mo, Yikuan Xia, Hao Zhu, Fangchen Liu, Minghua
  Liu, Hanxiao Jiang, Yifu Yuan, He~Wang, et~al.
\newblock Sapien: A simulated part-based interactive environment.
\newblock In \emph{Proceedings of the IEEE/CVF Conference on Computer Vision
  and Pattern Recognition}, pages 11097--11107, 2020.

\bibitem[Yi et~al.(2018)Yi, Huang, Liu, Kalogerakis, Su, and
  Guibas]{part_induction}
Li~Yi, Haibin Huang, Difan Liu, Evangelos Kalogerakis, Hao Su, and Leonidas
  Guibas.
\newblock Deep part induction from articulated object pairs.
\newblock \emph{arXiv preprint arXiv:1809.07417}, 2018.

\end{thebibliography}
\end{document}